\documentclass[letterpaper]{article} % DO NOT CHANGE THIS
\usepackage{aaai24}  % DO NOT CHANGE THIS
\usepackage{times}  % DO NOT CHANGE THIS
\usepackage{helvet}  % DO NOT CHANGE THIS
\usepackage{courier}  % DO NOT CHANGE THIS
\usepackage[hyphens]{url}  % DO NOT CHANGE THIS
\usepackage{natbib}
\usepackage{adjustbox}
\usepackage{tabularray}

\usepackage{graphicx} % DO NOT CHANGE THIS
\usepackage{natbib}  % DO NOT CHANGE THIS AND DO NOT ADD ANY OPTIONS TO IT
\usepackage{caption} % DO NOT CHANGE THIS AND DO NOT ADD ANY OPTIONS TO IT
\usepackage{booktabs}
\usepackage{multirow}

\usepackage{algorithm}
\usepackage{algorithmic}

\usepackage{newfloat}
\usepackage{listings}
\DeclareCaptionStyle{ruled}{labelfont=normalfont,labelsep=colon,strut=off} % DO NOT CHANGE THIS
\floatstyle{ruled}
\newfloat{listing}{tb}{lst}{}
\floatname{listing}{Listing}
\title{Continued Pretraining for Domain Adaptation of Wav2vec2.0 in Automatic Speech Recognition for Elementary Math Classroom Settings}
\author {
    Ahmed Adel Attia\textsuperscript{\rm 1},
    Dorottya Demszky\textsuperscript{\rm 2},
    Tol\'{u}l\d{o}p\d{\'{e}} \`{O}g\'{u}nr\d{\`{e}}m\'{i}\textsuperscript{\rm 2},
    Jing Liu\textsuperscript{\rm 1},
    Carol Espy-Wilson\textsuperscript{\rm 1}
    }
\affiliations {
    \textsuperscript{\rm 1}University of Maryland\\
    \textsuperscript{\rm 2}Stanford University\\
    aadel@umd.edu, 	
    ddemszky@stanford.edu, tolulope@cs.stanford.edu,
jliu28@umd.edu, espy@umd.edu	
}
\usepackage{bibentry}
\begin{document}

\maketitle

\begin{abstract}
Creating Automatic Speech Recognition (ASR) systems that are robust and resilient to classroom conditions is paramount to the development of AI tools to aid teachers and students. In this work, we study the efficacy of continued pretraining (CPT) in adapting Wav2vec2.0 to the classroom domain. We show that CPT is a powerful tool in that regard and reduces the Word Error Rate (WER) of Wav2vec2.0-based models by upwards of 10\%. More specifically, CPT improves the model's robustness to different noises, microphones, classroom conditions as well as classroom demographics. Our CPT models show improved ability to generalize to different demographics unseen in the labeled finetuning data.
\end{abstract}
\vspace{-15pt}

\section{Introduction}
Ensuring equitable access to high quality educational opportunities remains a persistent challenge in the US education system \cite{reardon2018widening,barrett2021disparities,xie2015stem,morrison2023critical}. Disparities in instruction quality, and specifically teacher-student interactions, contribute significantly to systemic inequities \cite{darling2004inequality}. Providing teachers with feedback to improve their instruction has the potential to increase teacher effectiveness and reduce inequities \citep{kraft2018effect,link2022grading}, but is highly resource intensive to implement at scale. AI has the potential to complement expert feedback and provide teachers with low-cost, consistent, automated feedback, which can improve their instruction and student outcomes \cite{demszky2023a,demszky2023improving,demszky2023mpowering,jacobs2022promoting}. Such tools can foster active learning environments where students can contribute and feel heard, which in turn can foster equity in the classroom. %NOTE: This is an anonymized manuscript. We can expand more on the M-Powering teachers project in the camera ready subission. 

Automatic Speech Recognition (ASR) is a critical component in the pipeline of tools needed to provide automated feedback. Transcripts generated by ASR systems can be analyzed on many levels to understand the dynamics in the classrooms --- if they are sufficiently accurate \citep{demszky2023improving,jacobs2022promoting,jacobs2024automated}. Recent advancements in transformer models have allowed ASR systems to witness a major boom and approach human-level performance in transcribing clean American English speech \cite{radford2023robust}. This level of performance is difficult to achieve for classroom ASR \cite{southwell2022challenges}. Classrooms are uniquely characterized by the abundance of children's speech and unique classroom noise, like children's babble noise and as well as other conditions that are known to affect the accuracy of ASR systems, like far-field speech \cite{zhu2024multichannel}, and multi-speaker conditions \cite{chang2019end, chang2020end}. 

\subsection{Challenges of Children ASR}
% and require extensive finetuning to achieve acceptable results \cite{attia2023kid, shahnawazuddin2024developing,jain2023adaptation, southwell2024automatic}.
 % Finetuning is necessary given 
ASR systems struggle with children's speech out-of-the-box, even in clean conditions. ASR systems are mainly trained with adult speech and, therefore cannot deal with the fact that children speak less clearly than adults \cite{lee1999acoustics} and that the basic acoustic and linguistic characteristics of children's speech differ from that of adults \cite{gerosa2009review}. Additionally, children's speech exhibits more inter-speaker variability due to varying developmental rates and intra-speaker variability due to underdeveloped pronunciation skills  \cite{koenig2008speech, koenig2008stop, lee1999acoustics, lee1997analysis, vorperian2007vowel,  smith1992relationships},  or immigration from non-English-speaking countries. In fact, in the United States, a significant percentage of students are classified as English Language Learners 18\% in California, 20\% in Texas, 10\% nationally \cite{nces2020}. %The NCTE corpus \cite{demszky2022ncte} which surveyed around 2000 4th and 5th grade elementary math classes from largely historically marginalized communities across several Northeastern states between 2010 and 2013, showed  21\% of the 10,817 students surveyed exhibited limited English proficiency, which affects ASR models.

In addition to the issues mentioned above, children's speech exhibits unique linguistic properties that ASR systems are not adapted to.  \citet{attia2023kid} did an analysis the performance of Whisper \cite{radford2023robust}, a popular transformer-based ASR system, on popular children's speech corpora and found that the language model in Whisper struggles with the fact that children change topics multiple times in the sentence in unstructured spontaneous speech. They found that Whisper can achieve adult-level performance with simple and scripted prompts, proving that Whisper can adapt to the acoustic properties of children's speech, but requires more work to adapt to the linguistic properties. Several other publications have shown that finetuning improves the performance of popular ASR systems, but a gap still exists between adult and children's speech  \cite{attia2023kid, shahnawazuddin2024developing,jain2023adaptation, southwell2024automatic}.
\subsection{Challenges of Classroom environments}
In classrooms, however, the previously mentioned challenges are compounded by multiple factors, namely the presence of unique noises and the presence of multiple speakers. 

In the US, classrooms hold around 20 students on average \cite{nces_census}. In collaborative learning scenarios, children's babble noise severely affects the performance of ASR. Babble noise, which is the noise from multiple speakers in the background, is considered one of the most challenging noises even in adult speech with adult babble \cite{simic2024self}. However, children babble noise is even more challenging, as this kind of noise is unlikely to exist in public datasets used to train these models.  The severity of noise increases with the number of students in the classroom and disproportionately affects overcrowded schools. 

Additionally, classrooms are a multi-speaker environment, where an unpredetermined number of unseen speakers speak in the same classroom, possibly overlapping with each other. Multi-speaker ASR is an open area of research in and of itself, and a wide gap still exists between single and multi-speaker ASR \cite{chang2019end, chang2020end}. In classrooms, there is an abundance of overlapping speech which complicates the multi-speaker ASR problem further. Far-field speech is another problem, for example, if a student asks a question while being far from the microphone. Far-field speech is another open area of research, and previous works have shown it to be a much more difficult task than near-field speech \cite{zhu2024multichannel}.

% as they have been shown to be sensitive to accented English in children's speech \cite{jain2024exploring}.
\subsection{Data Scarcity}
All of the previous challenges are compounded by the fact that transcribed classroom datasets are scarce, and those that exist are not public, partially due to the sensitivity of data representing minors. Although non-transcribed classroom recordings do exist, transcription can be prohibitively expensive. This low-resource setting lends itself well to self-supervised speech representation models like Wav2vec2.0 \cite{baevski2020wav2vec}, where the untranscribed data can be used to pretrain Wav2vec2.0 and the limited transcribed data can be used for finetuning for ASR. %Recent works \cite{hsu2021robust, san2024predicting, nowakowski2023adapting} have investigated continued pretraining (CPT) as an effective tool for domain adaptation. Robust Wav2vec2  has investigated the adaptation of a model pretrained on TED-LIUM v3 (TD) \cite{hernandez2018ted} dataset to LibriSpeech (LS) \cite{panayotov2015librispeech} and found that both joint training (pretraining on TD and LS together) and CPT (pretraining on TD then on LS) achieved similar performance. In the realm of low-resource languages,  one study has investigated adapting an already pretrained cross-lingual speech representation model, XLSR-53 \cite{conneau2020unsupervised}, to an unseen language by first performing CPT on 234 hours of untranscribed recordings, and then finetuning on 9 hours of that same language. Under that paradigm, they were able to achieve an absolute improvement in Word Error Rate (WER) of 7.5\% when compared to finetuning XLSR-53 directly. 

\subsection{How Continued Pretraining Can Help}
In this paper, we propose continued pretraining (CPT) as an effective way to adapt Wav2vec2.0 to domain-specific noisy speech recognition, namely classroom speech recognition. We consider three different 300M parameter variations of Wav2vec2.0:Wav2vec2.0 pretrained on LibriVox-60, XLS-R \cite{babu2021xls} pretrained on speech from 128 languages, and Robust Wav2vec2.0 pretrained on noisy English speech. We perform CPT on unlabeled classroom data for each model and then finetune the resultant models on labeled classroom data for ASR. In addition, we perform finetuning of the off-the-shelf Wav2vec2.0 models without CPT. We also pretrain Wav2vec2.0 from scratch on the same data used in CPT as it is the baseline used in similar publications \cite{zhu2022noise}. Our results show that CPT is the most effective way for adapting Wav2vec2.0 for noisy classroom speech and that the choice of starting point for CPT affects performance. In fact, our best-performing model outperforms Whisper \cite{radford2023robust}
when finetuned on the same dataset. 

\textbf{Our contributions} in this research can be summarized as follows:
\begin{itemize}
    \item We show that CPT is the most effective tool to adapt Wav2vec2.0 to noisy conditions like classrooms. 
    \item We show how CPT can improve the robustness of Wav2vec2.0, not only to noise but to different microphone configurations and demographics.
    \item Our proposed methods are more robust to noise and different demographics than Whisper, both off-the-shelf and finetuned.
    \item We provide a detailed analysis showing how ASR models might be biased against minority teachers and how to mitigate these problems moving forward.
    \item We demonstrate the use of existing classroom text corpora for LM pretraining.
\end{itemize}

To facilitate further research and reproducibility, our training code as well as our model checkpoints will be available at the time of the camera-ready submission.

\section{Background: Wav2vec2.0}

\begin{figure}[h]
    \centering
    \includegraphics{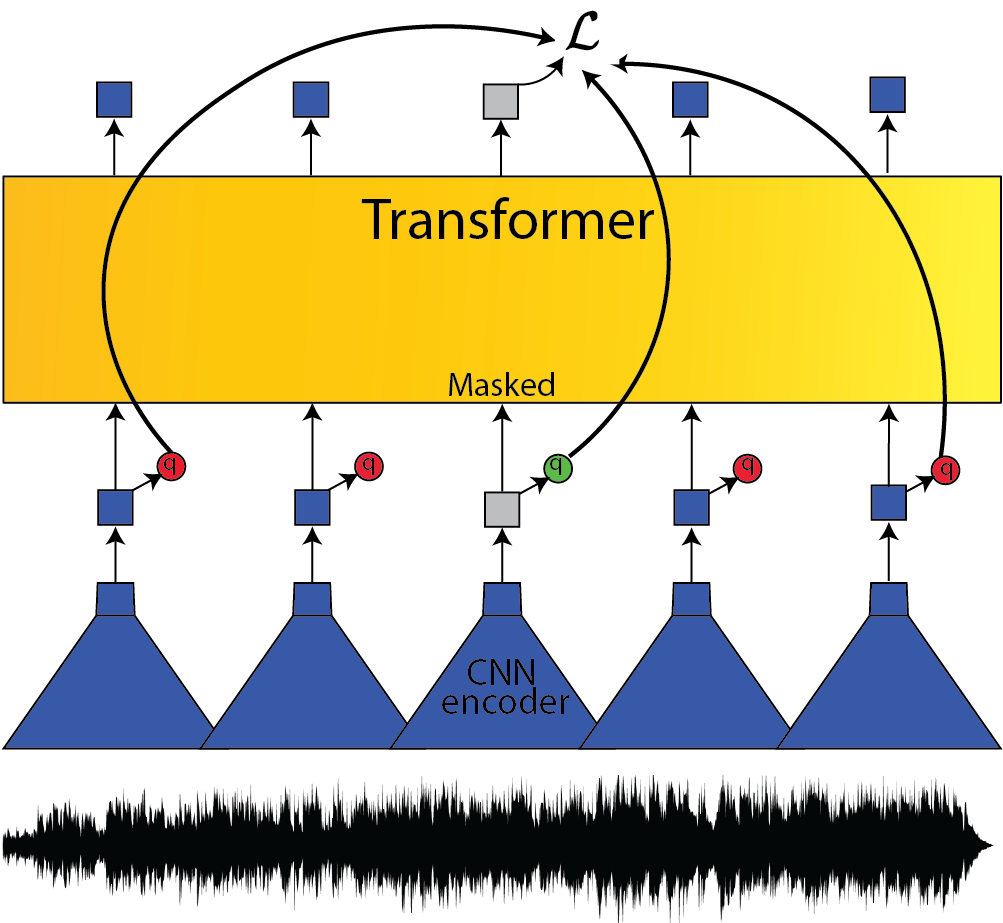}
    \caption{Wav2vec2.0 pretraining architecture. Lowercase q in circles represents the quantization networks, with the green circle representing the positive sample and the red circles representing the negative samples. Adapted from \cite{baevski2020wav2vec}}
    \label{fig1:wav2vec2}
\end{figure}

Wav2vec2.0 is a Self Supervised Learning (SSL) speech representation model developed by \citet{baevski2020wav2vec}, in a follow-up to Wav2vec \cite{schneider2019wav2vec}. Wav2vec2.0 utilizes the contextualization capabilities of transformers to learn contextual self-supervised speech representations from unlabeled audio in an SSL paradigm. 

Supervised speech models, like Whisper, learn directly on task-specific labeled data. That means they need a large amount of human-transcribed data to achieve SOTA performance. On the flip side, SSL models can learn useful speech representations from unlabeled data, either through non-contrastive learning, by extracting targets from the input speech signal to perform predictive learning, like in the case of HuBERT \cite{hsu2021hubert}, or through contrastive learning by contrasting positive and negative examples, like in Wav2vec. 

Wav2vec2.0's architecture consists of a convolutional feature extractor that extracts latent representations from raw audio and a transformer network that produces contextual representations of speech. During self-supervised pretraining, the latent representations generated by the convolutional feature encoder are quantized and the unquantized latent features are fed into the transformer, with some frames masked. The model learns through a constrastive learning task which is to distinguish the quantized representations corresponding to the masked frames from the context. Additionally, a diversity loss is also applied to diversify the quantized representations.

For ASR finetuning, a single randomly initialized layer is added on top of the transformer network, whose size corresponds to the number of characters in the vocabulary plus a single word boundary token. The model is then optimized using a  Connectionist Temporal Classification (CTC) loss.

\section{Related Previous Works}
In this section, we discuss previous works of interest related to noise robustness and domain adaptation in Wav2vec2.0
\subsection{Robust Wav2vec2.0}
\citet{hsu2021robust} investigated the effect of using target domain data during the pretraining. Their findings suggest that adding in-domain data during pretraining improves performance when the resulting model is finetuned on that in-domain data. Additionally, they also found that adding more data even if it is out-of-domain still improves performance, but less so than in-domain data. 

Most relevant to our work, are their experiments with CPT. They show that performing CPT with more unlabeled in-domain data and then finetuning on out-of-domain data improves the performance on an in-domain test set. These results serve as a strong motivation for our work, however, they do not sufficiently answer our question about the effectiveness of CPT for improving noise robustness for two main reasons: (1) they do not finetune on their in-domain dataset and, (2) their in-domain test dataset is Librispeech \textit{dev-other}, and while it is more challenging than \textit{dev-clean}, it is still considered a very clean dataset compared to the challenging domain of classroom ASR. 

\subsection{Noise-Robust Wav2vec2.0}
The work proposed by \citet{zhu2022noise} investigates the noise robustness of Wav2vec2.0 to noise and proposes a new pretraining paradigm that performs implicit speech enhancement on the latent representations in Wav2vec2.0. Simply put, during pretraining they feed a clean frame as well as a noise-augmented frame to the same copy of the convolutional encoder network and then force the output latent representations to be closer together via a consistency loss. They then feed the noisy latent representations to the transformer, but they sample the positive and negative samples from the quantized clean features.

Their experiments on vanilla-wav2vec2.0 show that pretraining on noisy speech improves the performance on both in-domain and out-of-domain finetuning ASR tasks when compared to clean pretraining. Their results show that their proposed pretraining structure achieves a 4-5\% absolute improvement in Word Error Rate (WER) in noisy speech across several signal-to-noise ratio (SNR) values as well as less degradation in clean speech performance. The main drawback of that method is that it requires the use of clean-noisy pairs for pretraining and doesn't generalize to naturally occurring noisy recordings like classroom recordings. However, their experiments show that even without any modifications, pretraining on noisy speech improves the performance in both in-domain and out-of-domain finetuning scenarios. 

\subsection{Adaptation of Wav2vec2.0 to low-resource languages through CPT}
Several research papers have attempted to investigate the effectiveness of CPT in adapting good multi-lingual self-supervised ASR systems like XLSR53 \cite{conneau2020unsupervised} and XLS-R \cite{babu2021xls}. The research by \cite{nowakowski2023adapting} aimed to develop an ASR system for Ainu, a critically endangered and low-resource language local to northern Japan, Sakhalin, and Kuril Islands. Starting from XLSR53 which was already pretrained on 56K hours from 53 languages, they performed CPT on 234 hours of Ainu recordings. Then they performed multiple finetuning experiments using different subsets of their labeled training data. They describe CPT as "clearly the most effective way to adapt a speech representation model for a new language". CPT decreased their WER by up to 40\% relative to the unadapted model, an absolute improvement of about 20.6\%.

Another interesting study in the field of low-resource languages \cite{san2024predicting} performed CPT on different languages, including the target language (Punjabi) as well as languages similar and different to it starting from XLS-R 128. Their best-performing setup was with CPT on the full 70 hours available of the target language (22.2\% WER) and they found better performance when supplementing the target language with a related language like Hindi (23.4\% WER) versus using unrelated languages like Malayalam, Bengali, Odia, or Tamil  (25\% WER), however, even CPT on a mixture of 10 hours of Punjabi and 60 hours of unrelated languages improved the performance relative to the unadapted XLS-R 128 model (30.8\% WER).

% Another interesting study in the field of low-resource language \cite{san2024predicting} investigated CPT when supplementing limited in-domain data with out-of-domain data. With a total of 70 hours of Punjabi, only 10 hours of which are used for finetuning, CPT on the full 70 hours starting from XLS-R 128 yields superior performance to the unadapted model. When replacing 60 hours of the CPT data with 60 hours from different languages, they found that similar languages yielded better performance than unrelated ones, but both cases yielded superior performance to the unadapted model.

\section{Datasets}
\subsection{Audio Datasets}
\subsubsection{NCTE}
The NCTE dataset consists of video and audio recordings of 2128 4th and 5th-grade elementary math classrooms collected as part of the National Center for Teacher Effectiveness (NCTE) Main Study \cite{kane2022}. The observations took place between 2010 and 2013 across four districts serving historically marginalized students. 

Table \ref{tab:NCTE} shows the key demographics of the populations of students and teachers in the recordings. The statistics show that the students are balanced by gender, but the vast majority of them come from minority racial backgrounds (72.3\%). The majority of the students received free or reduced-price lunches (64.9\%) indicating that they likely come from low-income households. A sizable percentage of the students exhibited limited English proficiency (19.7\%) or special educational status (12.3\%). On the other hand, the teachers were mostly female (82.4\%) and White (64.2\%). This disparity between students' and teachers' demographics is within the national statistics \cite{demszky2022ncte}.
\begin{table}[h]
    \centering
    \caption{Key demographics from the NCTE dataset. LEP indicates that the students have Limited English Proficiency, FRPL indicates that they receive Free or Reduced-Price Lunches in the given school year and SPED indicates Special Education status. }
    
\begin{adjustbox}{width=.8\columnwidth,center}        
    \begin{tabular}{c||c|c}
    \hline
    \rule{0pt}{2ex}Statistic & Teachers &  Students\\
 
    \hline \hline
    \rule{0pt}{2ex}Number & 313 & 12661\\
    \hline
    \rule{0pt}{2ex}\% Male & 16.3\% & 49.4\%\\
    \% Female & 82.4\% & 49.6\%\\
    \% No data & 1.3\% & 1\%\\
    \hline
    \rule{0pt}{2ex}\% African-American & 22.4\%& 42.3\%\\
    \% Asian &2.6\%& 7.4\% \\
    \% Hispanic & 2.9\% & 22.6\% \\
    \% White & 64.2\% & 22.7\% \\
    \% Other & 3.6\% & 3.9\% \\
    \% No data & 0\% & 1\% \\
    \hline
    \rule{0pt}{2ex}\% LEP & - & 19.7\%\\
    \% No data & - & 1.2\% \\
    \hline
   \rule{0pt}{2ex} \% SPED & - & 12.3\%\\
    \% No data & - & 1.2\% \\
    \hline
    \rule{0pt}{2ex}\% FRPL & - & 64.9\%\\
    \% No data & - & 1.2\%\\

    \hline
    \hline
    
    \end{tabular}
    \end{adjustbox}
    
    \label{tab:NCTE}
    
\end{table}

\begin{table*}[h]
 \centering
 \LARGE
\caption{Key statistics from the transcribed portion of the NCTE dataset. \% Teacher refers to the percentage of speech duration attributed to the teacher. AFAM refers to African-American. FF indicates a far-field microphone and NN indicates a near-field microphone.}
 
\begin{adjustbox}{width=2.2\columnwidth,center}
\begin{tabular}{c|cc||c|ccccc|c||c||c} 
         \hline
 
         & \multicolumn{2}{|c||}{\textbf{Teacher}} & \multicolumn{7}{|c||}{\textbf{Students}} & &\\
         \hline
      \multicolumn{11}{c}{\textbf{Used for training and validation}}\\
         \hline
          
\textbf{Class ID} & \textbf{Gender}  & \textbf{Race}  & \textbf{\% Male} & \textbf{\% AFAM} & \textbf{\% Asian} & \textbf{\% Hispanic} & \textbf{\% White} &\textbf{ \% Other} & \textbf{\# Students} & \textbf{\% Teacher
} & \textbf{Microphone} \\
         \hline
\textbf{144}      & Female  & White                   & 85\%    & 62\%           & 0\%             & 8\%            & 31\%            & 0\%      & 13   & 84\%    & NF      \\
\textbf{622}      & Female  & AFAM                   & 53\%    & 16\%           & 11\%            & 47\%           & 26\%            & 0\%       &19    &79\%    & NF           \\
\textbf{2619}     & Female  & White                   & 48\%    & 76\%           & 5\%             & 0\%            & 19\%            & 0\%      &21     & 84\%   & FF           \\
\textbf{2709}     & Female  & White                   & 63\%    & 13\%           & 29\%            & 13\%           & 42\%            & 4\%        &24    & 68\%    & NF         \\
\textbf{2944}     & Female  & White                   & 54\%    & 46\%           & 4\%             & 8\%            & 38\%            & 4\%     &26   & 84\%    & NF         \\
\textbf{4724}     & Female  & White                   & 46\%    & 46\%           & 0\%             & 11\%           & 43\%            & 0\%         &28    & 92\%     & NF       \\
         \hline
\textbf{Total}     & -  & -                   & 56\%    & 42\%           & 8\%           & 14\%             & 34\%             & 2\%              &131         & NF        \\
\hline \hline
       \multicolumn{11}{c}{\textbf{Used for testing and analysis}} \\
         \hline
\textbf{13}      & Male  & Asian                   & 61\%    & 9\%           & 65\%             & 13\%            & 9\%            & 4\%    & 23  & 78\% & NF     \\
% \textbf{230}      & Male  & Asian                   & 61\%    & 9\%           & 65\%             & 13\%            & 9\%            & 4\%  & NF      & 23  \\
\textbf{4106}     & Female  & AFAM                   & 50\%    & 14\%           & 50\%             & 21\%            & 0\%            & 14\%              &12   & 81\%   & NF           \\
\textbf{4352}     & Male  & White                   & 41\%    & 34\%           & 10\%            & 7\%           & 48\%            & 0\%   &29  & 88\%  & NF           \\

\textbf{4651}     & Female  & AFAM                   & 55\%    & 27\%           & 14\%             & 9\%            & 50\%            & 0\%               &22      & 80\%  & FF         \\
\hline
\textbf{Total}     & -  & -                   & 51\%    & 23\%           & 32\%           & 11\%             & 31\%             & 3\%    &88    & NF        \\
         \hline
         \hline

\end{tabular}
\end{adjustbox}
 \label{tab:ncte_trans}

\end{table*}

For each classroom, 2 to 3 video and audio recordings exist, from different angles and microphones, each lasting 45 minutes to an hour. The total duration of the recordings from all microphones and classrooms amounts to 5235 hours. We resampled the audio from the original 44.1KHz sampling rate to 16KHz and cut it into 20-second chunks. About 10\% of the data was reserved for validation, and the rest was used as unlabeled pretraining data.

Out of these recordings, 6 were initially randomly chosen to be transcribed to create a low-resource unbalanced problem. The duration of this subset is about 5.15 hours, with the duration of each file between 45-60 minutes. Key statistics about the transcribed recordings are in Table \ref{tab:ncte_trans}. All teachers are White women except for one who is an African-American woman. The student's gender is mostly balanced throughout the entire dataset, with about 56\% of the students being male. All classes but one have a 45-65\% male population. Class 144 has 85\% male students. The majority of the students in the dataset come from minority racial backgrounds (66\%), with African-American students making up 42\% of the total student population, and both Asian and Hispanic students making up 22\%. This racial imbalance in the dataset presents a unique challenge and an interesting question to be answered: will pretraining on a racially diverse dataset improve the fairness and decrease the racial bias of the ASR system?  All but one recording in the dataset was recorded by a near-field microphone and the recording of class 2619 comes from a far-field microphone. This distinction is by design, to test how well the ASR system generalizes to microphone configurations unseen in the training data and whether pretraining on different microphone set-ups improves this generalization. 
 
 An additional 4 classroom recordings were transcribed, and reserved for testing. In total, this subset is about 3.7 hours. The racial make-up of this subset is different than the one trained on, with white students representing roughly 31\% of the population, African-American students representing 23\%, and Asian students being the most represented group with 32\% while being the least represented group in the training-validation subset, making up 65\% of one class and 50\% of another. Additionally, this subset differs from the train-validation subset in the presence of two male teachers, one of them being Asian. We use this dataset to answer the question: how well does the ASR system generalize to different make-ups than the ones seen in training? Does pretraining on a diverse dataset improve this generalization?

\subsubsection{In house dataset}
We recorded 6 classrooms from 3  schools. We recorded two 8th-grade classes in a charter school in Washington, DC that serve predominantly African-American and Hispanic low-income students. Classes included at least one special education student and at least one English language learner. We refer to these recordings as \textbf{DC-1} and \textbf{DC-2}. Two 5th-grade classrooms were recorded in a school in Eastlake, Ohio, with predominantly White students. Classes included special education students. We refer to these recordings as \textbf{OH-1} and \textbf{OH-2}.  The remaining two recordings came from 6th-grade classrooms at a private school in San Jose, California with predominantly White and Asian high-income students with no special education or English language learner students. We refer to these recordings as \textbf{CA-1} and \textbf{CA-2}.  In each classroom, five microphones were placed at different places in the classroom and the audio streams were then added together. This resulted in a good capture of all the audio in the classroom but also extremely noisy audio in one particularly noisy classroom, CA-1. We keep this configuration to test the model's ability to handle extremely noisy conditions. We use this dataset to test the effect of CPT on improving performance in slightly different but related domains than the one seen during CPT.
\subsubsection{NCTE-Text}
\label{subsubsec:ncte-text}
\citet{demszky2022ncte} published a dataset of anonymized transcriptions of 1660 classrooms from the NCTE dataset for the purposes of developing computational measures of classroom discourse. These transcripts were not intended for ASR training and are not suitable for it, because all the teachers and students were anonymized to protect the subjects' privacy, replacing every name with its initial. Additionally, the transcripts are not verbatim and do not capture the exact speech in the classroom word for word. Add to that that they were not properly time-stamped, making preprocessing very complicated. However, this text corpus still holds tremendous value for ASR. 

One of the key differences between Wav2vec2.0 and Whisper is that Whisper has a built-in audio-conditional LM, while Wav2vec2.0 needs an external n-gram or transformer LM for beam-search. Even though end-to-end models are generally preferable, breaking down the ASR problem into an acoustic model (Wav2vec2.0) and an LM allows us to utilize unmappable representations like non-verbatim transcriptions better. In that sense, the NCTE-Text dataset was very useful for use in training a custom 5-gram language model for beam-search decoding with Wav2vec2.0 

The corpus is de-identified, with teacher and student names replaced with initials such as "Student K" and "Mrs. D". To make this data suitable for LM finetuning, we replaced the initials with randomly chosen names. To ensure that the names are unbiased towards race or gender, we referenced a list of the most popular baby names by race between 2011 and 2019 in New York City \footnote{\url{https://data.cityofnewyork.us/Health/Popular-Baby-Names/25th-nujf/data}}. For each classroom transcription and every initial, we uniformly sample gender, and then sampled a race uniformly from White, African-American, Hispanic, or Asian. After sampling the race and gender, we referred to the list of popular names of that gender and race based on its popularity according to the list. We then replaced all instances of that anonymized initial in that transcription. Race-aware deanonymization ensures that all races and both genders are equally represented in the text corpus. Our results show that race-aware deanonymization shows some improvement over naive gender-aware deanonymization which does not take race into account.

\section{Experiments}
\subsection{CPT Experiments}
We performed three CPT experiments in addition to contrast the effect of the starting checkpoint for CPT on domain adaptation. For the CPT experiments, we started from three 300M parameter checkpoints of Wav2vec2.0. The first was initially pretrained on 60K hours LibriVox \cite{kearns2014librivox}, which we denote as \textbf{W2V-LV60}. The second model, \textbf{W2V-Robust} \cite{hsu2021robust} was pretrained on LibriVox, as well as noisy English recordings. The third model is \textbf{XLS-R} 300M which was pretrained on 436K hours of speech from 128 languages, including English. The contrast between the performance of W2V-LV60 and W2V-Robust will showcase the effect of additional noisy data from a different domain during initial pretraining on the efficacy of CPT domain adaptation. The performance of XLS-R will showcase the impact of additional pretraining data from completely different domains and languages.  We also pretrain a Wav2vec2.0 model from scratch to establish a baseline, which we denote as \textbf{W2V-SCR}, representing the baseline configuration from previous works.

We implemented our model using the official fairseq library \footnote{\url{https://github.com/facebookresearch/fairseq}}. Each of the models was pretrained for 1M steps on 3 NVIDIA-A100 GPUs using the configuration file obtained from the fairseq repository. 

\subsection{Finetuning}

We perform two separate 6-fold cross-validation finetuning on each of the two datasets used for training, leaving one classroom recording for validation in each fold. For each fold, we train using a configuration file adapted from the 10-hour configuration file found on the fairseq repository. Each fold was trained on one NVIDIA-A6000 GPU using the fairseq library. We finetune each off-the-shelf Wav2vec2.0 model (W2V-LV60, W2V-Robust, and XLS-R), and their CPT counterparts as well as W2V-SCR. In total, we pretrain 4 models and finetune 84 models, 42 for NCTE and 42 for our in-house dataset. 

We preprocess our data by cutting the audio into chunks of 30 seconds or less depending on the timestamps of the transcription. We also normalize the text using the WhipserNormalizer Library \footnote{\url{https://pypi.org/project/whisper-normalizer/}}.

\subsection{N-Gram LM Training}
We train our 5-gram LM using the KenLM\footnote{\url{https://github.com/kpu/kenlm}} library. Our training data is the deanonymized NCTE-Text dataset as described above. We normalize the training data using the Whisper Normalizer library to match the transcription text. We use this LM for LM decoding of Wav2vec2.0 in all our experiments.
\begin{table}[h]
\centering
\caption{Average cross-validation WER finetuning results starting from different Wav2vec2.0 checkpoints with and without continued pretraining, as well as comparison with the small English-only checkpoint of Whisper. Whisper-FT refers to finetuning Whisper on the target dataset. The standard deviation of the WER is between brackets. All models decoded with an n-gram LM use our LM trained on race-aware deanonymized NCTE-Text corpus.}

\begin{adjustbox}{width=\columnwidth,center}\begin{tabular}{cccc}
\hline
                       &                      & \multicolumn{2}{c}{\textbf{WER(STD)}}                                                \\ \cline{3-4}
\rule{0pt}{2ex}\multirow{-2}{*}{\textbf{Model}} &\multirow{-2}{*}{\textbf{LM}} & \textbf{NCTE}                        & \textbf{In-house}                    \\ \hline\hline
\rule{0pt}{2ex}\textbf{Whisper}          & -          & 24.46(12.37) & 30.15(12.99)                         \\
\rule{0pt}{2ex}\textbf{Whisper-FT}       & -          & 19.14(6.77)        & 28.53(14.07) 
                                              \\ \hline\hline
\multicolumn{4}{c}{\rule{0pt}{3ex}\textbf{\textit{No Continued Pretraining}}} \\
\rule{0pt}{2ex}\multirow{2}{*}{\textbf{W2V-LV60K}}     & None & 39.11(13.01)       & 37.82(12.30)                         \\
      & 5-gram LM  & 30.39(14.48)       & 33.56(10.86)                         \\ \hline

\rule{0pt}{2ex}\multirow{2}{*}{\textbf{XLS-R}} & None          & 38.19(10.39)       & 39.12(13.60)                         \\
                                & 5-gram LM  & 29.02(10.96)       & 32.49(10.76)                         \\\hline

\rule{0pt}{2ex}\multirow{2}{*}{\textbf{W2V-Robust}}     & None & 35.07(11.85)       & 36.36(11.54)                          \\
     & 5-gram LM  & 27.99(13.28)       &           31.49(9.97)                           \\ \hline \hline
\multicolumn{4}{c}{\rule{0pt}{3ex}\textbf{\textit{Pretraining from Scratch}}} \\
     
\rule{0pt}{2ex}\multirow{2}{*}{\textbf{W2V-SCR}}   & None          &    47.34(5.73)    &  51.39 (6.83)                                     \\
 &  5-gram LM &      30.25(15.44)      &           38.59(12.93)                                               
                       \\ \hline \hline

\multicolumn{4}{c}{\rule{0pt}{3ex}\textbf{\textit{Continued Pretraining}}} \\
    \rule{0pt}{2ex} \multirow{2}{*}{\textbf{W2V-LV60K (CPT)}}    & None &            22.52(4.89)        &             32.26(8.92)                        \\
                              & 5-gram LM  &        18.13(5.50)            &          26.72(7.72)                            \\ \hline
    \rule{0pt}{2ex} \multirow{2}{*}{\textbf{XLS-R (CPT)}}                          & None & 26.53(5.13)        & 32.16(10.78)                         \\
      & 5-gram LM  & 19.37(4.91)        & 26.80(8.00)                          \\ \hline
\rule{0pt}{2ex}\multirow{2}{*}{\textbf{W2V-Robust (CPT)}}    & None & 25.04(5.28)        & 30.97(9.99)                          \\
    & 5-gram LM  & \textbf{17.71(5.06)}                 & \textbf{26.50(8.09)}                \\ \hline\hline
\end{tabular}%
\end{adjustbox}
\label{tab:results_main}
\end{table}

\begin{table}[h]
\centering
\caption{Detailed WER results on each fold of the cross-validation on both the NCTE and in-house datasets, for the off-the-shelf and the CPT version of W2V-Robust and the finetuned small English-only Whisper checkpoint.}

\begin{adjustbox}{width=\columnwidth,center}
\begin{tabular}{cccc}
\hline
\multicolumn{4}{c}{\rule{0pt}{2ex}\textbf{NCTE}}                                                      \\
\hline
\rule{0pt}{2ex}\textbf{Fold}    & \textbf{Whisper-FT} & \multicolumn{1}{c}{\textbf{W2V-Robust}} & \textbf{W2V-Robust (CPT)} \\\hline
\rule{0pt}{2ex}\textbf{144}  & \textbf{14.78}      &              19.86                        & 15.20                     \\
\rule{0pt}{2ex}\textbf{622}  & \textbf{16.97}      &              26.31                        & 18.77                     \\
\rule{0pt}{2ex}\textbf{2619} & 28.4                &              54.03                        & \textbf{27.15}            \\
\rule{0pt}{2ex}\textbf{2709} & \textbf{12.74}      &              19.09                        & 13.12                     \\
\rule{0pt}{2ex}\textbf{2944} & 26.98               &              28.07                        & \textbf{17.61}            \\
\rule{0pt}{2ex}\textbf{4724} & 14.95               &              20.55                        & \textbf{14.43}            \\
\hline
\rule{0pt}{2ex}\textbf{Average}       & 19.14               &          27.99                            & \textbf{17.71}            \\
\hline \hline
\multicolumn{4}{c}{\rule{0pt}{2ex}\textbf{In-House}}                                                  \\\hline
\rule{0pt}{2ex}\textbf{Fold}      & \textbf{Whisper-FT} & \multicolumn{1}{c}{\textbf{W2V-Robust}} & \textbf{W2V-Robust (CPT)} \\\hline
\rule{0pt}{2ex}\textbf{OH-1}                          & 19.76               &             18.55                         & \textbf{15.42}            \\
\rule{0pt}{2ex}\textbf{OH-2}                          & \textbf{16.42}      &             19.89                         & 16.88                     \\
\rule{0pt}{2ex}\textbf{DC-1}                          & 28.79               &             29.42                         & \textbf{24.90}            \\
\rule{0pt}{2ex}\textbf{DC-2}                          & \textbf{26.42}      &             37.32                         & 30.34                     \\
\rule{0pt}{2ex}\textbf{CA-1}                          & 55.77               &             49.03                         & \textbf{41.54}            \\
\rule{0pt}{2ex}\textbf{CA-2}                          & \textbf{24.04}      &             34.47                         & 29.91                     \\
\hline
\rule{0pt}{2ex}\textbf{Average}                           & 28.53               &         31.45                             & \textbf{26.50}      
\\\hline\hline
\end{tabular}%
\end{adjustbox}
\label{tab:detailed}
\end{table}

% \usepackage{graphicx}
\begin{table}[h]
\caption{Test set WER results on the held-out files from the NCTE dataset not used in cross-validation training. Each entry is the average WER of all the cross-validation versions of a particular model when tested on a particular recording in the test set, with the standard deviation in brackets. Average shows the total average WER of each model.}

\centering
\LARGE
\begin{adjustbox}{width=\columnwidth,center}
\begin{tabular}{ccccc}
\hline
\textbf{Model}     & \textbf{Whisper} & \textbf{Whisper-FT} & \textbf{W2V-Robust} & \textbf{W2V-Robust(CPT)} \\
\hline\hline
\textbf{13}     & 26.02            & 25.89(1.70)        & 25.55(0.66)                  & \textbf{25.07(0.73)}              \\
\textbf{4106} & \textbf{25.10}           & 35.65(8.17)         & 30.59(0.90)                 & 25.79(0.26)                       \\
\textbf{4352} &  22.27             & 22.13(0.41)         & 26.15(0.59)                  & \textbf{22.30(0.26)}              \\
\textbf{4651} & 50.74            & 39.82(6.88)         & 35.19(0.43)            & \textbf{29.47(0.40)}              \\
\hline
\textbf{Average}       & 31.03            & 31.12               & 30.12                        & \textbf{25.66}        \\           
\hline\hline
\end{tabular}%
\end{adjustbox}
\label{tab:results_test}
\end{table}
\section{Results and Discussion}

In this section, we showcase our finetuning results of different Wav2vec2.0 checkpoints as well as the results from the small English-only Whisper checkpoint both off-the-shelf and finetuned in the leave-one-out cross-validation fashion described in the \textbf{Experiments} section. 
\subsection{Overall performance}
Table \ref{tab:results_main} shows the average cross-validation WER across all folds for every model. Without CPT, finetuning W2V-Robust yields superior performance. This shows that initial pretraining on noisy data improves the noise robustness to unseen noises in different domains. Additionally, XLS-R yields slightly better results than W2V-LV60K, which shows that training on much more data from other languages can be useful, especially since some of these recordings might be noisier than LV-60K. LM decoding improves results by 6\% on average, which is in line with the results from the Appendix of the seminal Wav2vec2.0 paper \cite{baevski2020wav2vec}. There is a slight performance gap between NCTE and In-house indicating that the NCTE task is slightly easier, which is in line with the noise levels observed upon manual inspection of the recordings. The standard deviation in the results is quite high, which shows that each recording has unique characteristics and thus the folds perform differently on each one, showing that although we are considering classroom recordings to be one domain, the kind of classroom environment be it collaborative learning or purely instructional affects the noise level and the amount of adult teacher's speech versus students' speech which affects the performance. One thing to note is that one of the files comes from a far-field microphone to test how well the ASR system generalizes to unseen microphone configuration which we expand on in Table \ref{tab:detailed}.

Looking at the third and fourth sections of Table \ref{tab:results_main}, the advantage of CPT when compared to pretraining from scratch is immediately clear. Pretraining from scratch yields the worst performance in the entire table. Previous works \cite{zhu2022noise} have shown that pretraining on noisy data yields superior performance to pretraining on clean data, but in their tests, they pretrained both configurations on the same dataset but augmented the training data with noise for noisy pretraining. However, in our experiments, the clean pretraining data is at least 12 times bigger than the noisy pretraining data, which explains why clean self-supervised pretraining yields better performance in our experiments. One interesting thing to note is that the gap between the LM-decoded and the Viterbi-decoded results when pretraining from scratch is much higher than with other configurations. This suggests that the model does learn useful acoustical representations but does not learn enough linguistic properties during pretraining, which is accounted for by LM decoding. This result also suggests that initial pretraining on clean adult speech, even from other languages learns useful linguistic representations that are not sufficiently learned from noisy in-domain pretraining from scratch. Note the wide gap between the Viterbi decoded WER from W2V-LV60K(CPT) and W2V-SCR of almost 25\%.  In that regard, CPT has the benefit of utilizing linguistic representations from clean, and out-of-domain speech as well as learning the acoustic properties, and further learning useful linguistic properties from in-domain data.

Looking at CPT results in the fourth section of Table \ref{tab:results_main}, we can see that for NCTE with LM decoding, CPT improves WER by 10.73\% on average which proves that CPT is a powerful tool for domain adaptation when labeled data is scarce and unlabeled data is plentiful. For the In-house dataset where the domain is slightly different, CPT is still very effective, yielding a performance improvement of about 5\% on average, proving that CPT is effective in domain adaptation in a generalizable fashion. It is also noticeable how the standard deviation decreases in both datasets, down to 5.06 in NCTE and 8.09 in the In-house dataset. This shows that performing CPT on diverse classroom environments and noise conditions improves the ability of the model to generalize to these conditions. 

In terms of the choice of starting point for CPT, the order of performance with CPT does not follow the order observed in the performance without CPT. W2V-LV60K outperforms XLS-R with CPT, meaning that the performance edge XLS-R had from initially pretraining on more data from different languages does not carry forward with CPT, and initial pretraining on English-only datasets is always superior. This adds another dimension to the findings by \citet{hsu2021robust} which showed that adding more pretraining data, even if out-of-domain improves performance. Our findings suggest that this does not extend to CPT.  However, W2V-Robust still provides the best performance, showing that initial pretraining on noisy data even if from completely different domains yields better speech representations when adapted to other domains.

Finally, we can see that with CPT, W2V-Robust outperforms the Whisper checkpoint of comparable size, even with finetuning in both NCTE and In-house datasets. The nature of self-supervised speech models breaks down the problem into three parts: pretraining/CPT, finetuning, and LM decoding. This gives us more flexibility in situations when labeled data is scarce and expensive, but unlabeled speech data and text annotations from other domains are plentiful. Without CPT or LM decoding, Wav2vec2.0 performs much worse than Whisper, with WER being higher by 15-19\% in the NCTE dataset when finetuned on the same data in the same manner. However, the flexibility that Wav2vec2.0 allows beyond supervised finetuning improved the performance by up to 19\% through a combination of CPT and LM decoding. 

\subsection{Detailed analysis of cross-validation results}
In this section, we do a more detailed analysis of the results. We start by discussing the results in Table \ref{tab:detailed} which shows the WER of each fold of the cross-validation in Whisper-FT and W2V-Robust with and without CPT. 

\subsubsection{NCTE} 
Looking at the results, it is apparent that CPT significantly improves the performance in each fold of the cross-validation with one notable example recording 2619. Starting with recording 2619, this recording is the only one that comes from a far-field microphone, with the entirety of the training data in this cross-validation fold coming from near-field microphones. It is thus no surprise that without CPT, the model performs much worse in this fold than the others. However, with CPT, the error is cut in half, proving that CPT helps the model generalize to microphone configurations unseen in the labeled data but present in the unlabeled pretraining data. %Looking at recording 622, and by referencing Table \ref{tab:ncte_trans}, we can see that this is the only class in the training-validation dataset where the teacher is a African-American woman. We can see that without CPT, W2V-Robust underperforms on this file, and CPT improves the performance by 12\%. However, this class has the highest WER after 2619 which comes from a far-field microphone.

In terms of comparison with Whisper, looking at the NCTE results, CPT allows the model to achieve close performance or improve upon Whisper in every fold, with one major improvement noticed with recording 2944. Upon manual inspection, it was noted that this class started as an instructional class for 15 minutes and then the teacher assigned a set of questions to the students and started making rounds in the class. This resulted in a much noisier environment than other classrooms with a higher degree of children's babble noise. Whisper was unable to deal with children's babble noise, often interleaving the target speaker with whatever it could discern from the background noise and sometimes exhibiting characteristic Whisper hallucinations by repeating a single word or phrase, for example, \textit{``how do you know that what what is the denominator what is the denominator the the the the the..."} with the word \textit{"the"} repeating 206 times. Wav2vec does not suffer from the same hallucination problem, and with CPT, it's much more capable of dealing with children's babble noise and focusing on the target speaker correctly. 

%When we look at the performance of Whisper with class 622, we can again see that it has the highest WER after 2619 and the noisy 2944. Even though CPT allowed W2V-Robust to adapt to different noises well, the race of the teacher still seems to be a determining factor in the performance, which might indicate a lack of diversity in the dataset, however, we cannot say for sure since the there is not enough information about Whisper's training data.

\subsubsection{In-house dataset}
The differences in WER between Whisper and W2V-Robust with CPT on each fold are higher in the in-house dataset. W2V-Robust with CPT outperforms Whisper in 3 folds, with the main improvement coming from the recording of class CA-1. This recording is perhaps the noisiest of all the datasets used in this study, as discussed in the \textbf{Dataset} section. Both W2V-Robust and Whisper have high WER for this class recording, but even the non-CPT W2V-Robust outperformed Whisper. Upon manual inspection, we again see that Whisper suffers from extreme hallucinations with high children babble noise. On the other hand, W2V-Robust with CPT is more capable of handling extremely noisy conditions, outperforming Whisper in this fold by 14\%.

\subsection{Performance on an unseen test set with different demographics}
In this section, we discuss the results from the unseen NCTE test set. As previously discussed, and according to Table \ref{tab:ncte_trans}, the test set has a different racial makeup with better representations of Asian students and the presence of two male teachers, one of them being Asian, while all the teachers in the train-validation subset are White women except for one African-American woman. This test shows how well the models generalize to different demographics, unseen during training. 

From Table \ref{tab:results_test}, CPT improves the WER on average, from 30.12\% to 25.66\%, with the decreased standard deviation showing less variation in performance with different training data. This shows that CPT smoothes out the differences caused by different representations in the training data. When compared to Whisper, both the finetuned and non-finetuned versions of Whisper perform worse than both W2V-Robust models and finetuning seems to even harm the performance on average, indicating that Whisper overfits. 

However, it's important to note that even with CPT, the average WER with W2V-Robust is higher than the average cross-validation WER. Upon manual inspection, we noticed that the time stamps used to preprocess the audio into smaller chunks for testing had some inaccuracies. We attempted to account for that in our preprocessing, but we suspect that this issue still contributes to some extent to the gap in performance. However, our findings in comparing different models' performance on the test set still stand. We are currently in the process of refining these timestamps for more accurate analysis to accurately pinpoint the cause of the performance gap.

\section{Racial Bias in ASR systems}
Our analysis showed that ASR models might perform worse with minority teachers, in particular Asian and African American teachers. We note a pattern existing in both the validation and test results where the WER of classes with minority teachers is noticeably higher than other classes with similar or even higher noise levels and conditions. 

In the training-validation dataset, we note that the performance of class 622 is worse than similar classes with the same noise conditions and levels. According to Table \ref{tab:detailed}, comparing class 622 to classes 144, 2709, and 4724, all of which have similar acoustic properties, we see that all three models presented in this table have higher WER for this particular file, with the worst offender being W2V-Robust without CPT. Our proposed CPT does narrow the gap and improves the performance by 8\%, however the gap still exists. By referencing Table \ref{tab:ncte_trans}, we see that this class is the only class with an African-American teacher and has the highest percentage of Hispanic students and the second-lowest number of students in the dataset. 

Looking at the test set results in Table \ref{tab:results_test}, we see a similar pattern. The best performing class with all the models is class 4352 where the teacher is a White man. We note again the presence of some inaccuracies in the test-set transcriptions resulting from inaccurate timestamps, however, these inaccuracies are common and constant between all four classes, so the comparison between them still stands. Classes 4106 and 4651 score worse with all models, with the narrowest gap being with our W2V-Robust (CPT) model, about 10\%. By referencing Table \ref{tab:ncte_trans}, we see that class 4106 has a female African-American teacher, with the majority of the students being Asian. Class 4651 also had a female African-American teacher with a majority of White students. Class 13 also scores lower than class 4352, where the teacher is an Asian man and upon manual inspection, we note the presence of a light accent. However, his speech is perfectly enunciated and clear.  Upon further inspection of all the files, we find that these classes are not noisier than the 4352.  The teacher in class 4352 wore a lanyard microphone which interfered with his shirt buttons and caused a clicking sound throughout the recording. Class 4352 also has the highest number of students in class. All of these findings suggest that the ASR system should perform \textit{worse} and not \textit{better} in class 4352, but they don't. 

Our findings suggest that ASR systems, our proposed systems included, are biased against minority teachers. Our findings are in line with previous research \cite{martin2023bias, jain2024exploring} which indicates that ASR systems are biased against African-American Language (AAL), as well as accented speech. These findings highlight the urgent need for better representations in training datasets, as this disparity in performance has been attributed to the absence of recordings from diverse racial groups and dialects in popular speech corpora. For example, in the TIMIT \cite{garofolo1993timit} dataset, out of 630 speakers, only 4 are African-American and 538 are White. This in part explains the bias found in ASR models, although more research is needed.

\section{Conclusions and Future Work}
We have demonstrated how CPT is the best way to adapt Wav2vec2.0 models to different domains. By performing CPT on 5000 hours of classroom recordings, we've improved the model's ability to generalize to (1) different noise conditions, (2) different microphones, and (3) different demographics than those existing in the labeled training data. We've shown that CPT can improve the WER on noisy classroom data by 10\% on average and up to 27\% in specific conditions. Our results suggest that CPT should be the baseline for Wav2vec2.0 domain adaptation experiments, especially in noise robustness applications as it is far superior to pretraining from scratch.  We also propose a race-aware deanonymized classroom text dataset for LM training. 

There is an urgent need for more balanced and fair labeled classroom datasets. To that end, we are developing tools to sample recordings from the unlabeled NCTE dataset in a way to ensures balanced demographics and fair representation. We are also working towards a larger CPT trial, with an additional 15K hours of unlabeled classroom recordings. 

Lastly, we plan to expand on the work of \cite{zhu2022noise} of speech enhancement-based Wav2vec2.0 pretraining. We are working on simulating classroom noises to create a classroom noise dataset that can augment clean speech for use in such tools.

% \clearpage
\bibliography{main}
\clearpage
\appendix
\onecolumn
\section{Appendix: Full Results Table}
We provide the full cross-validation results table from the finetuning all the models mentioned in the manscript.
% \usepackage{graphicx}
\begin{table*}[h]
\centering
\caption{Full results from cross-validation finetuning off all finetuned Wav2vec2.0 models, both with and without CPT, as well as Whisper-small.en, both finetuned and not off-the-shelf.}

\begin{adjustbox}{width=1.15\columnwidth,center}
\begin{tabular}{c||c|c|||c|c||c|c||c|c||c|c||c|c||c|c||c|c}
\hline
\textbf{Fine-tuning data} & \multicolumn{16}{c}{\textbf{NCTE}}                                                                                                          
 \\ \hline                                                                                                                        
\textbf{Model}            & Whisper        & Whisper-FT     & \multicolumn{2}{c||}{W2V-SCR}   & \multicolumn{2}{c||}{W2V-LV60K} & \multicolumn{2}{c||}{W2V-LV60K (CPT)} & \multicolumn{2}{c||}{XLS-R}   & \multicolumn{2}{c||}{XLS-R (CPT)} & \multicolumn{2}{c||}{W2V-Robust} & \multicolumn{2}{c}{W2V-Robust (CPT)} 
\\ \hline 
\textbf{LM}               & \textbf{-}     & -              & None         & 5-g LM         & None          & 5-g LM        & None            & 5-g LM            & None         & 5-g LM       & None            & 5-g LM        & None           & 5-g LM        & None         & 5-g LM                
\\ \hline 
\textbf{144}              & 20.38          & \textbf{14.78} & 43.31        & \textbf{27.85} & 31.01         & 21.75         & 22.41           & 14.78             & 32.12        & 22.31        & 22.57           & 15.51         & 26.81          & 19.86         & 21.48        & 15.20                 \\
\textbf{622}              & 21.43          & \textbf{16.97} & 47.80        & \textbf{30.55} & 36.08         & 27.66         & 26.53           & 19.13             & 36.72        & 28.09        & 29.32           & 21.46         & 33.22          & 26.31         & 26.79        & 18.77                 \\
\textbf{2619}             & 47.6           & 28.4           & 57.75        & \textbf{42.86} & 62.86         & 58.48         & 34.58           & 28.39             & 57.77        & 50.50        & 35.47           & 28.04         & 57.58          & 54.03         & 34.58        & \textbf{27.15}        \\
\textbf{2709}             & 14.34          & \textbf{12.74} & 41.38        & 22.84          & 30.80         & 20.60         & 21.02           & 13.03             & 30.68        & 21.17        & 22.32           & 14.75         & 27.24          & 19.09         & 20.10        & 13.12                 \\
\textbf{2944}             & 27.98          & 26.98          & 48.20        & 26.64          & 45.02         & 32.22         & 25.65           & 18.05             & 40.99        & 28.56        & 26.20           & 19.31         & 37.87          & 28.07         & 25.17        & \textbf{17.61}        \\
\textbf{4724}             & 15             & 14.95          & 45.60        & 30.74          & 28.91         & 21.60         & 22.94           & 15.40             & 30.86        & 23.46        & 23.29           & 17.17         & 27.70          & 20.55         & 22.14        & \textbf{14.43}        \\
\textbf{Average}          & 24.46(12.37)   & 19.14(6.77)    & 47.34(5.73)  & 30.25(6.83)    & 39.11(13.01)  & 30.39(14.48)  & 25.52(4.89)     & 18.13(5.50)       & 38.19(10.39) & 29.02(10.96) & 26.53(5.13)     & 19.37(4.91)   & 35.07(11.85)   & 27.99(13.28)  & 25.04(5.28)  & \textbf{17.71(5.06)}  \\
\hline
\textbf{Fine-tuning data} & \multicolumn{16}{c}{\textbf{In-House}}                                                                                                      \\                                                                               \hline\hline                                                   
\textbf{OH-1}             & 27.65          & 19.76          & 33.60        & 23.34          & 23.25         & 20.40         & 18.75           & 15.84             & 24.34        & 20.01        & 19.81           & 16.68         & 21.77          & 18.55         & 18.61        & \textbf{15.42}        \\
\textbf{OH-2}             & 16.84          & \textbf{16.42} & 33.53        & 24.86          & 26.03         & 24.04         & 21.14           & 17.53             & 23.43        & 21.40        & 18.64           & 17.23         & 21.83          & 19.97         & 17.80        & 16.88                 \\
\textbf{DC-1}             & \textbf{23.73} & 28.79          & 52.01        & 37.70          & 35.99         & 32.43         & 31.33           & 24.69             & 39.31        & 32.50        & 30.18           & 24.61         & 35.63          & 29.44         & 30.62        & 24.90                 \\
\textbf{DC-2}             & 32.76          & \textbf{26.42} & 59.55        & 43.49          & 44.73         & 37.03         & 37.27           & 30.15             & 47.75        & 36.05        & 39.50           & 30.56         & 44.65          & 37.70         & 36.81        & 30.34                 \\
\textbf{CA-1}             & 54.44          & 55.77          & 72.88        & 57.33          & 56.45         & 50.99         & 48.33           & \textbf{40.69}    & 58.64        & 49.09        & 51.23           & 41.43         & 56.85          & 49.42         & 48.61        & 41.54                 \\
\textbf{CA-2}             & 25.48          & \textbf{24.04} & 56.74        & 44.80          & 40.44         & 36.46         & 36.74           & 31.54             & 41.26        & 35.86        & 33.59           & 30.30         & 37.43          & 33.84         & 33.40        & 29.91                 \\
\textbf{Averge}           & 30.15(12.99)   & 28.53(14.07)   & 51.39(15.44) & 38.59(12.93)   & 37.82(12.30)  & 33.56(10.86)  & 32.26(8.92)     & 26.74(7.72)       & 39.12(13.60) & 32.49(10.76) & 32.16(10.78)    & 26.80(8.00)   & 36.36(11.54)   & 31.49(9.74)   & 30.97(9.99)  & \textbf{26.50(8.09)}  \\ \hline

\end{tabular}%
\end{adjustbox}
\label{tab:my-table}
\end{table*}

\end{document}